\newif\ifdraft \draftfalse
\newif\iffull \fulltrue

\drafttrue\fullfalse

\documentclass[11pt]{article}
\usepackage{amsmath}

\def\epsilon{\varepsilon}

\usepackage{fullpage}
\usepackage{amsmath, amssymb, amsthm}
\usepackage{color}
\usepackage{xcolor}
\title{The Frontiers of Fairness in Machine Learning}
\author{Alexandra Chouldechova\thanks{Heinz College, Carnegie Mellon University. \texttt{achould@cmu.edu}} \and Aaron Roth\thanks{Department of Computer and Information Science, University of Pennsylvania. \texttt{aaroth@cis.upenn.edu}}}
\begin{document}
\maketitle

\begin{abstract}
The last few years have seen an explosion of academic and popular interest in algorithmic fairness. Despite this interest and the volume and velocity of work that has been produced recently, the fundamental science of fairness in machine learning is still in a nascent state. In March 2018, we convened a group of experts as part of a CCC visioning workshop to assess the state of the field, and distill the most promising research directions going forward. This report summarizes the findings of that workshop. Along the way, it surveys recent theoretical work in the field and points towards promising directions for research.
\end{abstract}

\section{Introduction}

The last decade has seen a vast increase both in the diversity of applications to which machine learning is applied, and to the import of those applications. Machine learning is no longer just the engine behind ad placements and spam filters: it is now used to filter loan applicants, deploy police officers, and inform bail and parole decisions, amongst other things. The result has been a major concern for the potential for data driven methods to introduce and perpetuate discriminatory practices, and to otherwise be \emph{unfair}. And this concern has not been without reason: a steady stream of empirical findings has shown that data driven methods can unintentionally both encode existing human biases and introduce new ones (see e.g. \cite{swee13,bolukbasi2016man,CBN17,gendershades} for notable examples).

At the same time, the last two years have seen an unprecedented explosion in interest from the academic community in studying fairness and machine learning. ``Fairness and transparency'' transformed from a niche topic with a trickle of papers produced every year (at least since the work of \cite{PRT08}) to a major subfield of machine learning, complete with a dedicated archival conference (ACM FAT*). But despite the volume and velocity of published work, our understanding of the \emph{fundamental questions} related to fairness and machine learning remain in its infancy. What should fairness mean? What are the \emph{causes} that introduce unfairness in machine learning? How best should we modify our algorithms to avoid unfairness? And what are the corresponding tradeoffs with which we must grapple?

In March 2018, we convened a group of about fifty experts in Philadelphia, drawn from academia, industry, and government, to asses the state of our understanding of the fundamentals of the nascent \emph{science} of fairness in machine learning, and to identify the unanswered questions that seem the most pressing. By necessity, the aim of the workshop was not to comprehensively cover the vast growing field, much of which is empirical. Instead, the focus was on theoretical work aimed at providing a scientific foundation for understanding algorithmic bias. This document captures several of the key ideas and directions discussed.  
\section{What We Know}
\subsection{Causes of Unfairness}
Even before we precisely specify what we mean by ``fairness'', we can identify common distortions that can lead off-the-shelf machine learning techniques to produce behavior that is intuitively unfair. These include:
\begin{enumerate}
\item \textbf{Bias Encoded in Data}: Often, the training data that we have on hand already includes human biases. For example, in the problem of recidivism prediction used to inform bail and parole decisions, the goal is to predict whether an inmate, if released, will go on to commit another crime within a fixed period of time. But we do not have data on who commits crimes --- we have data on who is arrested. There is reason to believe that arrest data --- especially for drug crimes --- is skewed towards minority populations that are policed at a higher rate \cite{rothwell2014war}. Of course, machine learning techniques are designed to fit the data, and so will naturally replicate any bias already present in the data.  There is no reason to expect them to \emph{remove} existing bias.
\item \textbf{Minimizing Average Error Fits Majority Populations}: Different populations of people have different distributions over features, and those features have different relationships to the label that we are trying to predict. As an example, consider the task of predicting college performance based on high school data. Suppose there is a majority population and a minority population. The majority population employs SAT tutors and takes the exam multiple times, reporting only the highest score. The minority population does not. We should naturally expect both that SAT scores are higher amongst the majority population, and that their relationship to college performance is differently calibrated compared to the minority population. But if we train a group-blind classifier to minimize overall error, if it cannot simultaneously fit both populations optimally, it will fit the majority population. This is because --- simply by virtue of their numbers --- the fit to the majority population is more important to overall error than the fit to the minority population. This leads to a different (and higher) distribution of errors in the minority population. This effect can be quantified, and can be partially alleviated via concerted data gathering efforts \cite{CJS18}.
\item \textbf{The Need to Explore}: In many important problems, including recidivism prediction and drug trials, the data fed into the prediction algorithm depends on the actions that algorithm has taken in the past. We only observe whether an inmate will recidivate \emph{if we release him}. We only observe the efficacy of a drug on patients to whom it is assigned. Learning theory tells us that in order to effectively learn in such scenarios, we need to \emph{explore} --- i.e. sometimes take actions we believe to be sub-optimal in order to gather more data. This leads to at least two distinct ethical questions. First, when are the individual costs of exploration borne disproportionately by a certain sub-population? Second, if in certain (e.g. medical) scenarios, we view it as immoral to take actions we believe to be sub-optimal for any particular patient, how much does this slow learning, and does this lead to other sorts of unfairness?
\end{enumerate}

\subsection{Definitions of Fairness}
With a few exceptions, the vast majority of work to date on fairness in machine learning has focused on the task of batch classification. At a high level, this literature has focused on two main families of definitions\footnote{There is also an emerging line of work that considers \emph{causal} notions of fairness (see e.g., \cite{kilbertus2017avoiding, kusner2017counterfactual, nabi2018fair}). We intentionally avoided discussions of this potentially important direction because it will be the subject of its own CCC visioning workshop.}: \emph{statistical} notions of fairness and \emph{individual} notions of fairness.  We briefly review what is known about these approaches to fairness, their advantages, and their shortcomings.

\subsubsection{Statistical Definitions of Fairness}
Most of the literature on fair classification focuses on \emph{statistical} definitions of fairness. This family of definitions fixes a
small number of protected demographic groups $\mathcal{G}$ (such as racial groups), and then ask for (approximate)
parity of some statistical measure across all of these groups. Popular measures include raw positive classification rate, considered in work such as \cite{CV10,KAS11,awareness,feldman2015certifying} (also sometimes known as \emph{statistical parity} \cite{awareness}), false positive and false negative rates \cite{Chou16,KMR16,HPS16,ZVGG17} (also sometimes known as equalized odds \cite{HPS16}), and positive predictive value \cite{Chou16,KMR16} (closely related to equalized calibration when working with real valued risk scores). There are others --- see e.g. \cite{berksurvey} for a more exhaustive enumeration. This family of fairness definitions is attractive because it is simple, and definitions from this family can be achieved without making any assumptions on the data and can be easily verified. However, statistical definitions of fairness do not on their own give meaningful guarantees to individuals or structured subgroups of the protected demographic groups. Instead they give guarantees to ``average'' members of the protected groups. (See \cite{awareness} for a litany of ways in which statistical parity and similar notions can fail to provide meaningful guarantees, and \cite{gerrymandering} for examples of how some of these weaknesses carry over to definitions which equalize false positive and negative rates.) Different statistical measures of fairness can be at odds with one another. For example, \cite{Chou16} and \cite{KMR16} prove a fundamental impossibility result: except in trivial settings, it is impossible to simultaneously equalize false positive rates, false negative rates, and positive predictive value across protected groups. Learning subject to statistical fairness constraints can also be computationally hard \cite{hardnesspaper}, although practical  algorithms of various sorts are known \cite{HPS16,ZVGG17,MSRpaper}.
\subsubsection{Individual Definitions of Fairness}
Individual notions of fairness, on the other hand, ask for constraints that bind on specific pairs of individuals, rather than on a quantity that is averaged over groups. For example,  \cite{awareness} give a definition which roughly corresponds to the constraint that ``similar individuals should be treated similarly'', where similarity is defined with respect to a task-specific metric that must be determined on a case by case basis. \cite{JKMR16} suggest a definition which roughly corresponds to ``less qualified individuals should not be favored over more qualified individuals'', where quality is defined with respect to the \emph{true} underlying label (unknown to the algorithm). However, although the semantics of these kinds of definitions can be more meaningful than statistical approaches to fairness, the major stumbling block is that they seem to require making significant assumptions. For example, the approach of \cite{awareness} pre-supposes the existence of an agreed upon similarity metric, whose definition would itself seemingly require solving a non-trivial problem in fairness, and the approach of \cite{JKMR16} seems to require strong assumptions on the functional form of the relationship between features and labels in order to be usefully put into practice. These obstacle are serious enough that it remains unclear whether individual notions of fairness can be made practical --- although attempting to bridge this gap is an important and ongoing research agenda.  
\section{Questions at the Research Frontier}

\subsection{Between Statistical and Individual Fairness}
Given the limitations of extant notions of fairness, is there a way to get some of the ``best of both worlds''? In other words, constraints that are practically implementable without the need for making strong assumptions on the data or the knowledge of the algorithm designer, but which nevertheless provide more meaningful guarantees to individuals? Two recent papers, \cite{gerrymandering} and \cite{multical} (see also \cite{gerrymandering2,multical2} for empirical evaluations of the algorithms proposed in these papers), attempt to do this by asking for statistical fairness definitions to hold not just on a small number of protected groups, but on an exponential or infinite class of groups defined by some \emph{class of functions} of bounded complexity. This approach seems promising: because ultimately they are asking for statistical notions of fairness, the approaches proposed by these papers enjoy the benefits of statistical fairness: that no assumptions need be made about the data, nor is any external knowledge (like a fairness metric) needed.  It also better addresses concerns about ``intersectionality'', a term used to describe how different kinds of discrimination can compound and interact for individuals who fall at the intersection of several protected classes.

At the same time, the approach raises a number of additional questions: what function classes are reasonable, and once one is decided upon (e.g. conjunctions of protected attribures) what features should be ``protected''? Should these only be attributes that are sensitive on their own, like race and gender, or might attributes that are innocuous on their own correspond to groups we wish to protect once we consider their intersection with protected attributes (for example clothing styles intersected with race or gender)? Finally, this family of approaches significantly mitigates  some of the weaknesses of statistical notions of fairness by asking for the constraints to hold on average not just over a small number of coarsely defined groups, but over very finely defined groups as well. Ultimately, however, it inherits the weaknesses of statistical fairness as well, just on a more limited scale.

Another recent line of work aims to weaken the strongest assumption needed for the notion of individual fairness from \cite{awareness}: namely that the algorithm designer has perfect knowledge of a ``fairness metric''. \cite{KRR18} assume that the algorithm has access to an oracle which can return an unbiased estimator for the distance between two randomly drawn individuals according to an unknown fairness metric, and show how to use this to ensure a statistical notion of fairness related to \cite{gerrymandering,multical} which informally states that ``on average, individuals in two groups should be treated similarly if on average the individuals in the two groups are similar'' --- and this can be achieved with respect to an exponentially or infinitely large set of groups. Similarly, \cite{GJKR18} assumes the existence of an oracle which can identify fairness violations when they are made in an online setting, but cannot quantify the extent of the violation (with respect to the unknown metric). It is shown that when the metric is from a specific learnable family, this kind of feedback is sufficient to obtain an optimal regret bound to the best fair classifier while having only a bounded number of violations of the fairness metric. \cite{RY18} consider the case in which the metric is known, and show that a PAC-inspired approximate variant of metric fairness generalizes to new data drawn from the same underlying distribution. Ultimately, however, these approaches all assume that fairness is perfectly defined with respect to some metric, and that there is some sort of direct access to it. Can these approaches be generalized to a more ``agnostic'' setting, in which fairness feedback is given by human beings who may not be responding in a way that is consistent with any metric?

\subsection{Data Evolution and Dynamics of Fairness}
The vast majority of work in \emph{computer science} on algorithmic fairness has focused on one-shot classification tasks. But real algorithmic systems consist of many different components that are combined together, and operate in complex environments that are dynamically changing, sometimes because of the actions of the learning algorithm itself. For the field to progress, we need to understand the dynamics of fairness in more complex systems.

Perhaps the simplest aspect of dynamics that remains poorly understood is how and when components that may individually satisfy notions of fairness compose into larger constructs that still satisfy fairness guarantees. For example, if the bidders in an advertising auction individually are fair with respect to their bidding decisions, when will the allocation of advertisements be ``fair'', and when will it not? \cite{pipelines} and \cite{DI18} have made a preliminary foray in this direction. These papers embark on a systematic study of fairness under composition, and find that often the composition of multiple fair components will not satisfy any fairness constraint at all. Similarly, the individual components of a ``fair'' system may appear to be unfair in isolation. There are certain special settings, e.g. the ``filtering pipeline'' scenario of \cite{pipelines} --- modeling a scenario in which a job applicant is selected only if she is selected at every stage of the pipeline --- in which (multiplicative approximations of) statistical fairness notions compose in a well behaved way. But the high level message from these works is that our current notions of fairness compose poorly. Experience from differential privacy \cite{DMNS06,DR14} suggests that graceful degradation under \emph{composition} is key to designing complicated algorithms satisfying desirable statistical properties, because it allows algorithm design and analysis to be \emph{modular}. Thus, it seems important to find satisfying fairness definitions and richer frameworks that behave well under composition.

In dealing with socio-technical systems, it is also important to understand how algorithms dynamically effect their environment, and the incentives of human actors. For example, if the bar (for e.g. college admission) is lowered for a group of individuals, this might increase the average qualifications for this group over time because of at least two effects: a larger proportion of children in the next generation grow up in households with college educated parents (and the opportunities this provides), and the fact that a college education is achievable can incentivize effort to prepare academically. These kinds of effects are not considered when considering either statistical or individual notions of fairness in one-shot learning settings. The economics literature on affirmative action has long considered such effects --- although not with the specifics of machine learning in mind: see e.g. \cite{FV92,coates,Beck10}. More recently, there have been some preliminary attempts to model these kinds of effects in machine learning settings --- e.g. by modeling the environment as a markov decision process \cite{JJKMR17}, considering the equilibrium effects of imposing statistical definitions of fairness in a model of a labor market \cite{HC18}, specifying the functional relationship between classification outcomes and quality \cite{delayed}, or by considering the effect of a classifier on a downstream Bayesian decision maker \cite{KRZ18}. However, the specific predictions of most of the models of this sort are brittle to the specific modeling assumptions made --- they point to the need to consider long term dynamics, but do not provide robust guidance for how to navigate them. More work is needed here.

Finally, decision making is often distributed between a large number of actors who share different goals and do not necessarily coordinate. In settings like this, in which we do not have direct control over the decision making process, it is important to think about how to \emph{incentivize} rational agents to behave in a way that we view as fair. \cite{incentives} takes a preliminary stab at this task, showing how to incentivize a particular notion of individual fairness in a simple, stylized setting, using small monetary payments. But how should this work for other notions of fairness, and in more complex settings? Can this be done by controlling the flow of information, rather than by making monetary payments (monetary payments might be distasteful in various fairness-relevant settings)? More work is needed here as well. Finally, \cite{cd17} take a welfare maximization view of fairness in classification, and characterize the cost of imposing additional statistical fairness constraints as well. But this is done in a static environment. How would the conclusions change under a dynamic model?
\subsection{Modeling and Correcting Bias in the Data}

Fairness concerns typically surface precisely in settings where the available training data is already contaminated by bias.  The data itself is often a product of social and historical process that operated to the disadvantage of certain groups.   When trained in such data, off-the-shelf machine learning techniques may reproduce, reinforce, and potentially exacerbate existing biases.  Understanding how bias arises in the data, and how to correct for it, are fundamental challenges in the study of fairness in machine learning.

\cite{bolukbasi2016man} demonstrate how machine learning can reproduce biases in their analysis of the popular word2vec embedding trained on a corpus of Google News texts (parallel effects were independently discovered by \cite{CBN17}).  The authors show that the trained embedding exhibit female/male gender stereotypes, learning that ``doctor'' is more similar to man than to woman, along with analogies such as ``man is to computer programmer as woman is to homemaker''.  Even if such learned associations accurately reflect patterns in the source text corpus, their use in automated systems may exacerbate existing biases.   For instance, it might result in male applicants being ranked more highly than equally qualified female applicants in queries related to jobs that the embedding identifies as male-associated.

Similar risks arise whenever there is potential for feedback loops.  These are situations where the trained machine learning model informs decisions that then affect the data collected for future iterations of the training process.  \cite{lum2016predict} demonstrate how feedback loops might arise in predictive policing if arrest data were used to train the model\footnote{Predictive policing models are generally proprietary, and so it is not clear whether arrest data is used to train the model in any deployed system.}.  In a nutshell, since police are likely to make more arrests in more heavily policed areas, using arrest data to predict crime hotspots will disproportionately concentrate policing efforts on already over-policed communities.  Expanding on this analysis, \cite{ensign2018runawayfeedbackloops} find that incorporating community-driven data such as crime reporting helps to attenuate the biasing feedback effects. The authors also propose a strategy for accounting for feedback by adjusting arrest counts for policing intensity.  The success of the mitigation strategy of course depends on how well the simple theoretical model reflects the true relationships between crime intensity, policing, and arrests.  Problematically, such relationships are often  unknown, and are very difficult to infer from data.   This situation is by no means specific to predictive policing.

Correcting for data bias generally seems to require knowledge of how the measurement process is biased, or judgments about properties the data would satisfy in an ``unbiased'' world.  \cite{friedler2016possibility} formalize this as a disconnect between the \emph{observed space}---features that are observed in the data, such as SAT scores---and the unobservable \emph{construct space}---features that form the desired basis for decision making, such as intelligence.  Within this framework, data correction efforts attempt to undo the effects of biasing mechanisms that drive discrepancies between these spaces.  To the extent that the biasing mechanism cannot be inferred empirically, any correction effort must make explicit its underlying assumptions about this mechanism.  What precisely is being assumed about the construct space? When can the mapping between the construct space and the observed space be learned and inverted? What form of fairness does the correction promote, and at what cost?  The costs are often immediately realized, whereas the benefits are less tangible.  We will directly observe reductions in prediction accuracy, but any gains hinge on a belief that the observed world is not one we should seek to replicate accurately in the first place.  This is an area where tools from causality may offer a principled approach for drawing valid inference with respect to unobserved counterfactually `fair' worlds.





\subsection{Fair Representations}

Fair representation learning is a data de-biasing process that produces transformations (\textit{intermediate representations}) of the original data that retain as much of the task-relevant information as possible while removing information about sensitive or protected attributes.
This is one approach to transforming biased observational data in which group membership may be inferred from other features, to a construct space where protected attributes are statistically independent of other features.
First introduced in the work of \cite{zemel2013learning}, fair representation learning produces a de-biased data set that may in principle be used by other parties without any risk of disparate outcomes.  \cite{feldman2015certifying} and \cite{mcnamara2017provably} formalize this idea by showing how the disparate impact of a decision rule is bounded in terms of its balanced error rate as a predictor of the sensitive attribute.

Several recent papers have introduced new approaches for constructing fair representations.   \cite{feldman2015certifying} propose rank-preserving procedures for \textit{repairing} features to reduce or remove pairwise dependence with the protected attribute.  \cite{johndrow2017algorithm} build upon this work, introducing a likelihood-based approach that can additionally handle continuous protected attributes, discrete features, and which promotes joint independence between the transformed features and the protected attributes.  There is also a growing literature on using adversarial learning to achieve group fairness in the form of statistical parity or false positive/false negative rate balance \cite{edwards2015censoring,beutel2017data,zhang2018mitigating,madras2018learning}.

Existing theory shows that the fairness promoting benefits of fair representation learning rely critically on the extent to which existing associations between the transformed features and the protected characteristics are removed.  Adversarial downstream users may be able to recover protected attribute information if their models are more powerful than those used initially to obfuscate the data.  This presents a challenge both to the generators of fair representations as well as to auditors and regulators tasked with certifying that the resulting data is fair for use.
More work is needed to understand the implications of fair representation learning for promoting fairness in the real world.




\subsection{Beyond Classification}
Although the majority of the work on fairness in machine learning focuses on batch classification, batch classification is only one aspect of how machine learning is used. Much of machine learning --- e.g. online learning, bandit learning, and reinforcement learning --- focuses on \emph{dynamic} settings in which the actions of the algorithm feed back into the data it observes. These dynamic settings capture many problems for which fairness is a concern. For example, lending, criminal recidivism prediction, and sequential drug trials all are so-called \emph{bandit} learning problems, in which the algorithm cannot observe data corresponding to counterfactuals. We cannot see whether someone not granted a loan \emph{would have} paid it back. We cannot see whether an inmate not released on parole \emph{would have} gone on to commit another crime. We cannot see how a patient \emph{would have} responded to a different drug.

The theory of learning in bandit settings is well understood, and it is characterized by a need to trade off \emph{exploration} with \emph{exploitation}. Rather than always making a myopically optimal decision, when counter-factuals cannot be observed, it is necessary for algorithms to sometimes take actions that appear to be sub-optimal so as to gather more data. But in settings in which decisions correspond to individuals, this means sacrificing the well-being of a particular person for the potential benefit of future individuals. This can sometimes be unethical, and a source of unfairness \cite{explore}. Several recent papers explore this issue. For example, \cite{BBK17} and \cite{KMRWW18} give conditions under which linear learners need not explore at all in bandit settings, thereby allowing for best-effort service to each arriving individual, obviating the tension between ethical treatment of individuals and learning. \cite{RSVW18} show that the costs associated with exploration can be unfairly bourn by a structured sub-population, and that counter-intuitively, those costs can actually increase when they are included with a majority population, even though more data increases the rate of learning overall. However, these results are all preliminary: they are restricted to settings in which the learner is learning a linear policy, and \emph{the data really is governed by a linear model}. While illustrative, more work is needed to understand real-world learning in online settings, and the ethics of exploration.

There is also some work on fairness in machine learning in other settings --- for example, ranking \cite{YS17,CSV17}, selection \cite{selection,KR18}, personalization \cite{celis2017fair}, bandit learning \cite{infinitebandit,calibratedbandit}, human-classifier hybrid decision systems \cite{defer}, and reinforcement learning \cite{JJKMR17,DTB17}. But outside of classification, the literature is relatively sparse. This should be rectified, because there are interesting and important fairness issues that arise in other settings --- especially when there are combinatorial constraints on the set of individuals that can be selected for a task, or when there is a temporal aspect to learning.

\subsection*{Acknowledgements}
This material is based upon work supposed by the National Science Foundation under Grant No. 1136993.  Any opinions, findings, and conclusions or recommendations expressed in this material are those of the authors and do not necessarily reflect the views of the National Science Foundation.

We are indebted to all of the participants of the CCC visioning workshop, held March 18-19 2018 in Philadelphia. The workshop discussion shaped every aspect of this document. We are grateful to Helen Wright and Ann Drobnis, who are instrumental in making the workshop happen. Finally, we are thankful to Cynthia Dwork, Sampath Kannan, Michael Kearns, Toni Pitassi, and Suresh Venkatasubramanian who provided valuable feedback on this report.
 \bibliographystyle{alpha}
\bibliography{refs}

\newcommand{\etalchar}[1]{$^{#1}$}
\begin{thebibliography}{KNRW18b}

\bibitem[ABD{\etalchar{+}}18]{MSRpaper}
Alekh Agarwal, Alina Beygelzimer, Miroslav Dud{\'\i}k, John Langford, and Hanna
  Wallach.
\newblock A reductions approach to fair classification.
\newblock In {\em Proceedings of the 35th International Conference on Machine
  Learning, {ICML}}, volume~80 of {\em {JMLR} Workshop and Conference
  Proceedings}, pages 2569--2577. JMLR.org, 2018.

\bibitem[BBC{\etalchar{+}}16]{explore}
Sarah Bird, Solon Barocas, Kate Crawford, Fernando Diaz, and Hanna Wallach.
\newblock Exploring or exploiting? social and ethical implications of
  autonomous experimentation in ai.
\newblock 2016.

\bibitem[BBK17]{BBK17}
Hamsa Bastani, Mohsen Bayati, and Khashayar Khosravi.
\newblock Exploiting the natural exploration in contextual bandits.
\newblock {\em arXiv preprint arXiv:1704.09011}, 2017.

\bibitem[BCZ{\etalchar{+}}16]{bolukbasi2016man}
Tolga Bolukbasi, Kai-Wei Chang, James~Y Zou, Venkatesh Saligrama, and Adam~T
  Kalai.
\newblock Man is to computer programmer as woman is to homemaker? debiasing
  word embeddings.
\newblock In {\em Advances in Neural Information Processing Systems}, pages
  4349--4357, 2016.

\bibitem[BCZC17]{beutel2017data}
Alex Beutel, Jilin Chen, Zhe Zhao, and Ed~H Chi.
\newblock Data decisions and theoretical implications when adversarially
  learning fair representations.
\newblock {\em arXiv preprint arXiv:1707.00075}, 2017.

\bibitem[Bec10]{Beck10}
Gary~S Becker.
\newblock {\em The economics of discrimination}.
\newblock University of Chicago press, 2010.

\bibitem[BG18]{gendershades}
Joy Buolamwini and Timnit Gebru.
\newblock Gender shades: Intersectional accuracy disparities in commercial
  gender classification.
\newblock In {\em Conference on Fairness, Accountability and Transparency},
  pages 77--91, 2018.

\bibitem[BHJ{\etalchar{+}}18]{berksurvey}
Richard Berk, Hoda Heidari, Shahin Jabbari, Michael Kearns, and Aaron Roth.
\newblock Fairness in criminal justice risk assessments: The state of the art.
\newblock {\em Sociological Methods \& Research}, 0(0):0049124118782533, 2018.

\bibitem[BKN{\etalchar{+}}17]{pipelines}
Amanda Bower, Sarah~N Kitchen, Laura Niss, Martin~J Strauss, Alexander Vargas,
  and Suresh Venkatasubramanian.
\newblock Fair pipelines.
\newblock {\em arXiv preprint arXiv:1707.00391}, 2017.

\bibitem[CBN17]{CBN17}
Aylin Caliskan, Joanna~J Bryson, and Arvind Narayanan.
\newblock Semantics derived automatically from language corpora contain
  human-like biases.
\newblock {\em Science}, 356(6334):183--186, 2017.

\bibitem[CDPF{\etalchar{+}}17]{cd17}
Sam Corbett-Davies, Emma Pierson, Avi Feller, Sharad Goel, and Aziz Huq.
\newblock Algorithmic decision making and the cost of fairness.
\newblock In {\em Proceedings of the 23rd ACM SIGKDD International Conference
  on Knowledge Discovery and Data Mining}, pages 797--806. ACM, 2017.

\bibitem[Cho17]{Chou16}
Alexandra Chouldechova.
\newblock Fair prediction with disparate impact: A study of bias in recidivism
  prediction instruments.
\newblock {\em Big data}, 5(2):153--163, 2017.

\bibitem[CJS18]{CJS18}
Irene Chen, Fredrik~D Johansson, and David Sontag.
\newblock Why is my classifier discriminatory?
\newblock 2018.

\bibitem[CL93]{coates}
Stephen Coate and Glenn~C Loury.
\newblock Will affirmative-action policies eliminate negative stereotypes?
\newblock {\em The American Economic Review}, pages 1220--1240, 1993.

\bibitem[CSV17]{CSV17}
L~Elisa Celis, Damian Straszak, and Nisheeth~K Vishnoi.
\newblock Ranking with fairness constraints.
\newblock {\em arXiv preprint arXiv:1704.06840}, 2017.

\bibitem[CV10]{CV10}
Toon Calders and Sicco Verwer.
\newblock Three naive bayes approaches for discrimination-free classification.
\newblock {\em Data Mining and Knowledge Discovery}, 21(2):277--292, 2010.

\bibitem[CV17]{celis2017fair}
L~Elisa Celis and Nisheeth~K Vishnoi.
\newblock Fair personalization.
\newblock {\em arXiv preprint arXiv:1707.02260}, 2017.

\bibitem[DHP{\etalchar{+}}12]{awareness}
Cynthia Dwork, Moritz Hardt, Toniann Pitassi, Omer Reingold, and Richard Zemel.
\newblock Fairness through awareness.
\newblock In {\em Proceedings of the 3rd innovations in theoretical computer
  science conference}, pages 214--226. ACM, 2012.

\bibitem[DI18]{DI18}
Cynthia Dwork and Christina Ilvento.
\newblock Fairness under composition.
\newblock {\em Manuscript}, 2018.

\bibitem[DMNS06]{DMNS06}
Cynthia Dwork, Frank McSherry, Kobbi Nissim, and Adam Smith.
\newblock Calibrating noise to sensitivity in private data analysis.
\newblock In {\em Theory of Cryptography Conference}, pages 265--284. Springer,
  2006.

\bibitem[DR14]{DR14}
Cynthia Dwork and Aaron Roth.
\newblock The algorithmic foundations of differential privacy.
\newblock {\em Foundations and Trends{\textregistered} in Theoretical Computer
  Science}, 9(3--4):211--407, 2014.

\bibitem[DTB17]{DTB17}
Shayan Doroudi, Philip~S. Thomas, and Emma Brunskill.
\newblock Importance sampling for fair policy selection.
\newblock In {\em Proceedings of the Thirty-Third Conference on Uncertainty in
  Artificial Intelligence, {UAI}}. {AUAI} Press, 2017.

\bibitem[EFN{\etalchar{+}}18]{ensign2018runawayfeedbackloops}
Danielle Ensign, Sorelle~A. Friedler, Scott Neville, Carlos Scheidegger, and
  Suresh Venkatasubramanian.
\newblock Runaway feedback loops in predictive policing.
\newblock In {\em 1st Conference on Fairness, Accountability and Transparency
  in Computer Science ({FAT*})}, 2018.

\bibitem[ES15]{edwards2015censoring}
Harrison Edwards and Amos Storkey.
\newblock Censoring representations with an adversary.
\newblock {\em arXiv preprint arXiv:1511.05897}, 2015.

\bibitem[FFM{\etalchar{+}}15]{feldman2015certifying}
Michael Feldman, Sorelle~A Friedler, John Moeller, Carlos Scheidegger, and
  Suresh Venkatasubramanian.
\newblock Certifying and removing disparate impact.
\newblock In {\em KDD}, 2015.

\bibitem[FSV16]{friedler2016possibility}
Sorelle~A Friedler, Carlos Scheidegger, and Suresh Venkatasubramanian.
\newblock On the (im) possibility of fairness.
\newblock {\em arXiv preprint arXiv:1609.07236}, 2016.

\bibitem[FV92]{FV92}
Dean~P Foster and Rakesh~V Vohra.
\newblock An economic argument for affirmative action.
\newblock {\em Rationality and Society}, 4(2):176--188, 1992.

\bibitem[GJKR18]{GJKR18}
Stephen Gillen, Christopher Jung, Michael Kearns, and Aaron Roth.
\newblock Online learning with an unknown fairness metric.
\newblock In {\em Advances in Neural Information Processing Systems}, 2018.

\bibitem[HC18]{HC18}
Lily Hu and Yiling Chen.
\newblock A short-term intervention for long-term fairness in the labor market.
\newblock In Pierre{-}Antoine Champin, Fabien~L. Gandon, Mounia Lalmas, and
  Panagiotis~G. Ipeirotis, editors, {\em Proceedings of the 2018 World Wide Web
  Conference on World Wide Web, {WWW}}, pages 1389--1398. {ACM}, 2018.

\bibitem[HJKRR18]{multical}
Ursula H{\'e}bert-Johnson, Michael~P Kim, Omer Reingold, and Guy~N Rothblum.
\newblock Calibration for the (computationally-identifiable) masses.
\newblock In {\em Proceedings of the 35th International Conference on Machine
  Learning, {ICML}}, volume~80 of {\em {JMLR} Workshop and Conference
  Proceedings}, pages 2569--2577. JMLR.org, 2018.

\bibitem[HPS16]{HPS16}
Moritz Hardt, Eric Price, and Nati Srebro.
\newblock Equality of opportunity in supervised learning.
\newblock In {\em Advances in neural information processing systems}, pages
  3315--3323, 2016.

\bibitem[JJK{\etalchar{+}}17]{JJKMR17}
Shahin Jabbari, Matthew Joseph, Michael Kearns, Jamie Morgenstern, and Aaron
  Roth.
\newblock Fairness in reinforcement learning.
\newblock In {\em International Conference on Machine Learning}, pages
  1617--1626, 2017.

\bibitem[JKM{\etalchar{+}}18]{infinitebandit}
Matthew Joseph, Michael Kearns, Jamie Morgenstern, Seth Neel, and Aaron Roth.
\newblock Fair algorithms for infinite and contextual bandits.
\newblock In {\em AAAI/ACM Conference on AI, Ethics, and Society}, 2018.

\bibitem[JKMR16]{JKMR16}
Matthew Joseph, Michael Kearns, Jamie~H Morgenstern, and Aaron Roth.
\newblock Fairness in learning: Classic and contextual bandits.
\newblock In {\em Advances in Neural Information Processing Systems}, pages
  325--333, 2016.

\bibitem[JL17]{johndrow2017algorithm}
James~E Johndrow and Kristian Lum.
\newblock An algorithm for removing sensitive information: application to
  race-independent recidivism prediction.
\newblock {\em arXiv preprint arXiv:1703.04957}, 2017.

\bibitem[KAS11]{KAS11}
Toshihiro Kamishima, Shotaro Akaho, and Jun Sakuma.
\newblock Fairness-aware learning through regularization approach.
\newblock In {\em Data Mining Workshops (ICDMW), 2011 IEEE 11th International
  Conference on}, pages 643--650. IEEE, 2011.

\bibitem[KCP{\etalchar{+}}17]{kilbertus2017avoiding}
Niki Kilbertus, Mateo~Rojas Carulla, Giambattista Parascandolo, Moritz Hardt,
  Dominik Janzing, and Bernhard Sch{\"o}lkopf.
\newblock Avoiding discrimination through causal reasoning.
\newblock In {\em Advances in Neural Information Processing Systems}, pages
  656--666, 2017.

\bibitem[KGZ18]{multical2}
Michael~P Kim, Amirata Ghorbani, and James Zou.
\newblock Multiaccuracy: Black-box post-processing for fairness in
  classification.
\newblock {\em arXiv preprint arXiv:1805.12317}, 2018.

\bibitem[KKM{\etalchar{+}}17]{incentives}
Sampath Kannan, Michael Kearns, Jamie Morgenstern, Mallesh Pai, Aaron Roth,
  Rakesh Vohra, and Zhiwei~Steven Wu.
\newblock Fairness incentives for myopic agents.
\newblock In {\em Proceedings of the 2017 ACM Conference on Economics and
  Computation}, pages 369--386. ACM, 2017.

\bibitem[KLRS17]{kusner2017counterfactual}
Matt~J Kusner, Joshua Loftus, Chris Russell, and Ricardo Silva.
\newblock Counterfactual fairness.
\newblock In {\em Advances in Neural Information Processing Systems}, pages
  4069--4079, 2017.

\bibitem[KMR17]{KMR16}
Jon~M. Kleinberg, Sendhil Mullainathan, and Manish Raghavan.
\newblock Inherent trade-offs in the fair determination of risk scores.
\newblock In {\em 8th Innovations in Theoretical Computer Science Conference,
  {ITCS}}, 2017.

\bibitem[KMR{\etalchar{+}}18]{KMRWW18}
Sampath Kannan, Jamie Morgenstern, Aaron Roth, Bo~Waggoner, and Zhiwei~Steven
  Wu.
\newblock A smoothed analysis of the greedy algorithm for the linear contextual
  bandit problem.
\newblock In {\em Advances in Neural Information Processing Systems}, 2018.

\bibitem[KNRW18a]{gerrymandering2}
Michael Kearns, Seth Neel, Aaron Roth, and Zhiwei~Steven Wu.
\newblock An empirical study of rich subgroup fairness for machine learning.
\newblock {\em arXiv preprint arXiv:1808.08166}, 2018.

\bibitem[KNRW18b]{gerrymandering}
Michael~J. Kearns, Seth Neel, Aaron Roth, and Zhiwei~Steven Wu.
\newblock Preventing fairness gerrymandering: Auditing and learning for
  subgroup fairness.
\newblock In Jennifer~G. Dy and Andreas Krause, editors, {\em Proceedings of
  the 35th International Conference on Machine Learning, {ICML}}, volume~80 of
  {\em {JMLR} Workshop and Conference Proceedings}, pages 2569--2577. JMLR.org,
  2018.

\bibitem[KR18]{KR18}
Jon Kleinberg and Manish Raghavan.
\newblock Selection problems in the presence of implicit bias.
\newblock {\em arXiv preprint arXiv:1801.03533}, 2018.

\bibitem[KRR18]{KRR18}
Michael~P Kim, Omer Reingold, and Guy~N Rothblum.
\newblock Fairness through computationally-bounded awareness.
\newblock In {\em Advances in Neural Information Processing Systems}, 2018.

\bibitem[KRW17]{selection}
Michael Kearns, Aaron Roth, and Zhiwei~Steven Wu.
\newblock Meritocratic fairness for cross-population selection.
\newblock In {\em International Conference on Machine Learning}, pages
  1828--1836, 2017.

\bibitem[KRZ18]{KRZ18}
Sampath Kannan, Aaron Roth, and Juba Ziani.
\newblock Downstream effects of affirmative action.
\newblock {\em arXiv preprint arXiv:1808.09004}, 2018.

\bibitem[LDR{\etalchar{+}}18]{delayed}
Lydia~T Liu, Sarah Dean, Esther Rolf, Max Simchowitz, and Moritz Hardt.
\newblock Delayed impact of fair machine learning.
\newblock In {\em Proceedings of the 35th International Conference on Machine
  Learning, {ICML}}, 2018.

\bibitem[LI16]{lum2016predict}
Kristian Lum and William Isaac.
\newblock To predict and serve?
\newblock {\em Significance}, 13(5):14--19, 2016.

\bibitem[LRD{\etalchar{+}}17]{calibratedbandit}
Yang Liu, Goran Radanovic, Christos Dimitrakakis, Debmalya Mandal, and David~C
  Parkes.
\newblock Calibrated fairness in bandits.
\newblock {\em arXiv preprint arXiv:1707.01875}, 2017.

\bibitem[MCPZ18]{madras2018learning}
David Madras, Elliot Creager, Toniann Pitassi, and Richard Zemel.
\newblock Learning adversarially fair and transferable representations.
\newblock {\em arXiv preprint arXiv:1802.06309}, 2018.

\bibitem[MOW17]{mcnamara2017provably}
Daniel McNamara, Cheng~Soon Ong, and Robert~C Williamson.
\newblock Provably fair representations.
\newblock {\em arXiv preprint arXiv:1710.04394}, 2017.

\bibitem[MPZ17]{defer}
David Madras, Toniann Pitassi, and Richard~S. Zemel.
\newblock Predict responsibly: Increasing fairness by learning to defer.
\newblock {\em CoRR}, abs/1711.06664, 2017.

\bibitem[NS18]{nabi2018fair}
Razieh Nabi and Ilya Shpitser.
\newblock Fair inference on outcomes.
\newblock In {\em Proceedings of the... AAAI Conference on Artificial
  Intelligence. AAAI Conference on Artificial Intelligence}, volume 2018, page
  1931. NIH Public Access, 2018.

\bibitem[PRT08]{PRT08}
Dino Pedreshi, Salvatore Ruggieri, and Franco Turini.
\newblock Discrimination-aware data mining.
\newblock In {\em Proceedings of the 14th ACM SIGKDD international conference
  on Knowledge discovery and data mining}, pages 560--568. ACM, 2008.

\bibitem[Rot14]{rothwell2014war}
Jonathan Rothwell.
\newblock How the war on drugs damages black social mobility.
\newblock {\em The Brookings Institution, published Sept}, 30, 2014.

\bibitem[RSVW18]{RSVW18}
Manish Raghavan, Alexandrs Slivkins, Jennifer~Wortman Vaughan, and
  Zhiwei~Steven Wu.
\newblock The unfair externalities of exploration.
\newblock In {\em Conference on Learning Theory}, 2018.

\bibitem[RY18]{RY18}
Guy~N Rothblum and Gal Yona.
\newblock Probably approximately metric-fair learning.
\newblock In {\em Proceedings of the 35th International Conference on Machine
  Learning, {ICML}}, volume~80 of {\em {JMLR} Workshop and Conference
  Proceedings}, pages 2569--2577. JMLR.org, 2018.

\bibitem[Swe13]{swee13}
Latanya Sweeney.
\newblock Discrimination in online ad delivery.
\newblock {\em Queue}, 11(3):10, 2013.

\bibitem[WGOS17]{hardnesspaper}
Blake Woodworth, Suriya Gunasekar, Mesrob~I Ohannessian, and Nathan Srebro.
\newblock Learning non-discriminatory predictors.
\newblock In {\em Conference on Learning Theory}, pages 1920--1953, 2017.

\bibitem[YS17]{YS17}
Ke~Yang and Julia Stoyanovich.
\newblock Measuring fairness in ranked outputs.
\newblock In {\em Proceedings of the 29th International Conference on
  Scientific and Statistical Database Management}, page~22. ACM, 2017.

\bibitem[ZLM18]{zhang2018mitigating}
Brian~Hu Zhang, Blake Lemoine, and Margaret Mitchell.
\newblock Mitigating unwanted biases with adversarial learning.
\newblock 2018.

\bibitem[ZVGG17]{ZVGG17}
Muhammad~Bilal Zafar, Isabel Valera, Manuel Gomez{-}Rodriguez, and Krishna~P.
  Gummadi.
\newblock Fairness beyond disparate treatment {\&} disparate impact: Learning
  classification without disparate mistreatment.
\newblock In {\em Proceedings of the 26th International Conference on World
  Wide Web, {WWW}}, pages 1171--1180. {ACM}, 2017.

\bibitem[ZWS{\etalchar{+}}13]{zemel2013learning}
Rich Zemel, Yu~Wu, Kevin Swersky, Toni Pitassi, and Cynthia Dwork.
\newblock Learning fair representations.
\newblock In {\em ICML}, 2013.

\end{thebibliography}
\end{document}